\documentclass[runningheads]{llncs}
\usepackage{esvect}
\usepackage{booktabs}
\usepackage{multirow}
\usepackage[T1]{fontenc}
\usepackage{pifont}
\newcommand{\cmark}{\ding{51}} 
\newcommand{\xmark}{\ding{55}} 
\usepackage{makecell}  
\usepackage{multirow}
\usepackage[table]{xcolor}
\definecolor{sigAAA}{HTML}{A5D6A7}  
\definecolor{sigAA}{HTML}{C8E6C9}   
\definecolor{sigA}{HTML}{E8F5E9}    
\usepackage{graphicx,verbatim,subcaption}
\usepackage{enumitem}
\usepackage{amsfonts}
\usepackage{amsmath}
\usepackage{pifont}
\usepackage{marvosym}
\usepackage[labelfont=bf, labelsep=period]{caption}

\usepackage[colorlinks, linkcolor=blue, anchorcolor=blue, citecolor=blue]{hyperref}

\newcommand{\corrmark}{\textsuperscript{\Letter}}

\begin{document}
%
\title{Prior-Anchored Debiasing for Long-Tailed Multi-Organ Pathology Report Generation}
\titlerunning{PriOrGen}
%

\author{
Feng Yang\inst{1} \and
Jie Liu\inst{1} \and
Yubo Pang\inst{1} \and
Peilin Chen\inst{1} \and
Xinheng Lyu\inst{2} \and\\
Shiqi Wang\inst{1}\corrmark \and
Howard Leung\inst{1}\corrmark \and
Ping Chen\inst{3}
}
\authorrunning{F. Yang et al.}
\institute{
City University of Hong Kong, Hong Kong SAR \and
University of Nottingham, United Kingdom \and
University of Massachusetts Boston, United States\\
\email{
fenyang6-c@my.cityu.edu.hk,
shiqwang@cityu.edu.hk,
ping.chen@umb.edu
}
}
  
\maketitle              
\begin{abstract}
Automated pathology report generation from Whole Slide Images (WSIs) has attracted increasing attention in digital pathology. However, existing methods are predominantly developed under single-organ settings, overlooking the multi-organ scenarios encountered in clinical practice, where organ types typically follow a long-tailed distribution. To address this gap, we identify two critical biases: (1) visual representation bias, where the encoder favors head-class patterns over tail-class discriminative features, and (2) textual decoding bias, where the decoder overfits to head-class narrative patterns, yielding diagnostically unreliable outputs for tail-class organs. To mitigate these two biases, we propose a novel \textbf{Pri}or-anchored multi-\textbf{Or}gan pathology report \textbf{Gen}eration framework (PriOrGen). Specifically, a Visual-Prototype Anchored Bottleneck module leverages the information bottleneck principle with learnable anchor representations to selectively retain diagnostically relevant visual information while filtering out head-biased redundancy. Secondly, a Meta-Report Anchored Bank module constructs an organ-specific meta-report anchored bank and retrieves organ-faithful textual priors to steer the decoder away from head-class narrative patterns. Extensive experiments on a multi-organ pathology dataset demonstrate that our method effectively mitigates long-tail biases and achieves superior report generation performance across both head and tail organ categories compared to state-of-the-art methods. Code is publicly available at \url{https://github.com/yangfeng0216/PriOrGen}.

\keywords{Whole Slide Image \and Pathology Report Generation \and Long-Tail Learning.}

\end{abstract}
\section{Introduction}


In the rapidly evolving field of digital pathology, the automated analysis of Whole Slide Images (WSIs) has become a pivotal area of research \cite{zongshu1,CLAM,zongshu3}. Among its various applications, pathology report generation aims to directly translate gigapixel-level visual patterns into structured, diagnostic-aware natural language. However, existing pathology report generation methods \cite{wsicaption,histgen,tran2024generating,BiGen} have been predominantly designed and validated under \textit{single organ} settings, where models are trained and evaluated on WSIs from the specific organ.

\begin{figure}[t] 
    \centering
    \includegraphics[width=\linewidth]{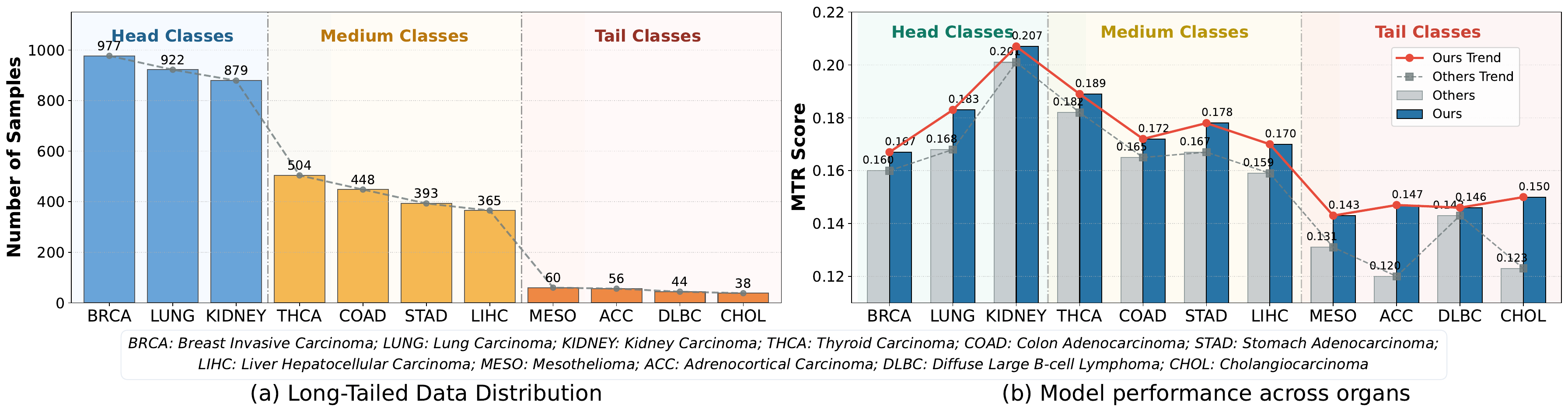} 
    \caption{\textbf{(a) Long-tailed data distribution}. The sample count per organ type exhibits a long-tailed distribution grouped into head, medium, and tail classes. \textbf{(b) Model performance across organs}. We compare the MTR score between our method and others.}
    \label{fig1}
\end{figure}

In real-world clinical scenarios, WSIs are derived from \textit{multiple organs} rather than a \textit{single organ}. To address this gap, this work investigates pathology report generation in a multi-organ setting, where organ types follow a long-tailed distribution (Fig.~\ref{fig1}(a)). Recent efforts have tackled long-tailed distributions in medical image classification~\cite{Long-Tai-cls1,Long-Tai-cls2,Long-Tai-cls3,Long-Tai-cls4}, where the output space is a fixed set of categorical labels. However, the impact of long-tailed distributions on pathology report generation remains largely unexplored. Unlike classification, report generation requires producing free-form texts that faithfully capture organ-specific characteristics, which renders the direct adoption of existing long-tail classification strategies insufficient.

Specifically, this introduces two biases into the pathology report generation pipeline. First, \textbf{visual representation bias}: when trained on long-tail data, the visual encoder tends to learn feature representations that are predominantly aligned with the histomorphological patterns of head-class organs~\cite{longtail1,longtail3}. Consequently, the discriminative visual cues of tail-class organs, which often exhibit distinct yet underrepresented morphologies, are inadequately captured, leading to degraded visual grounding for downstream report generation. Second, \textbf{textual decoding bias}: the language decoder, exposed to an overwhelming volume of head-class reports during training, tends to overfit to their dominant narrative patterns, including recurring sentence templates, high-frequency diagnostic phrases, and organ-specific terminologies~\cite{r2gen,longtail2}. As a result, when generating reports for tail-class organs, the decoder is prone to producing linguistically fluent but diagnostically erroneous outputs that implicitly borrow the stylistic and semantic patterns from head classes, undermining clinical reliability.

To address the above challenges, our method introduces two key components: (1) We propose a \textbf{Visual-Prototype Anchored Bottleneck} module that leverages the information bottleneck principle with learnable anchor representations to selectively filter out redundant or head-biased visual information, retaining diagnostically relevant patches from each WSI for more balanced and clinically grounded visual representations. (2) We propose a \textbf{Meta-Report Anchored Bank} module that constructs an organ-specific meta-report anchored bank by distilling representative report templates from the training reports of each organ type, and retrieves the most relevant meta-reports to provide the decoder with organ-faithful textual priors, steering the generation away from head-class narrative patterns. The main contributions of this paper can be summarized as follows:

\begin{itemize}[label=$\bullet$,leftmargin=*, itemsep=0pt, topsep=0pt, parsep=1pt]
  \item To the best of our knowledge, we propose the first framework that explicitly addresses the long-tail distribution bias in multi-organ pathology report generation, which manifests in two critical aspects: visual representation bias and textual decoding bias.
  
  

  \item We propose a dual prior mechanism consisting of a Visual-Prototype Anchored Bottleneck (VPAB) and a Meta-Report Anchored Bank (MRAB), which jointly mitigate long-tail bias from two perspectives. 
  
  \item Extensive experiments on the constructed ML-Path dataset demonstrate that our method achieves significant improvements over existing baselines, particularly on tail-class organs.
  
\end{itemize}

\section{Method}

\subsection{Overview}

\begin{figure}[t] 
    \centering
    \includegraphics[width=\linewidth]{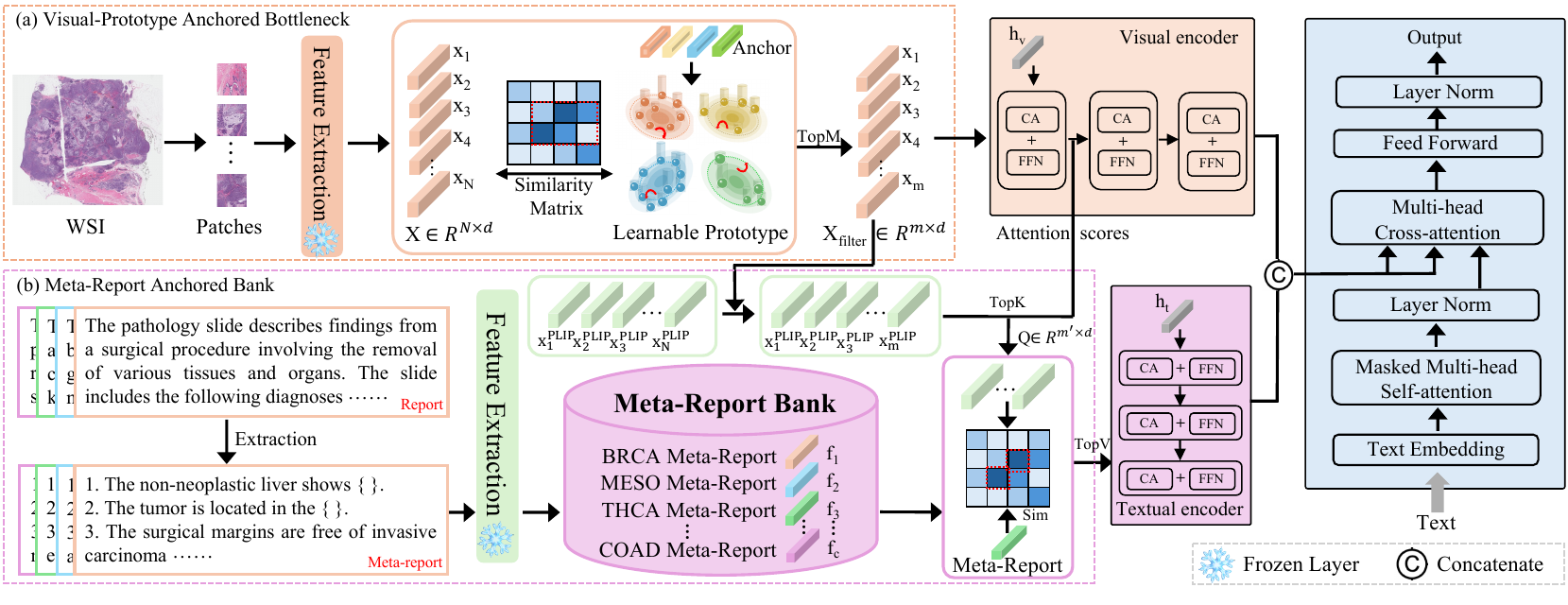} 
    \caption{\textbf{Overview of our proposed model PriOrGen.} CA and FFN refer to Cross-Attention and Feed-Forward Network. (a) Visual-Prototype Anchored Bottleneck module. (b) Meta-Report Anchored Bank module.}
    \label{fig2}
\end{figure}

As illustrated in Fig.~\ref{fig2}, our framework consists of three key components. Given patch features extracted from a WSI, the Visual-Prototype Anchored Bottleneck (VPAB, Sec.~\ref{sec:vib}) performs 
anchor-guided compression to selectively retain diagnostically relevant patches while suppressing head-biased redundancy. The Meta-Report Anchored Bank (MRAB, Sec.~\ref{sec:MRAB}) retrieves organ-faithful textual priors from a pre-constructed meta-report anchored bank to mitigate textual decoding bias. Finally, a cross-modal decoder (Sec.~\ref{sec:decoder}) integrates the compressed visual representations and retrieved textual 
priors to generate diagnostic reports.

\subsection{Problem Formulation}

Following standard practice in computational pathology~\cite{wsicaption,BiGen}, each WSI $\mathcal{W}$ is partitioned into $N$ non-overlapping patches at 10$\times$ magnification. Each patch is encoded by a pre-trained pathology foundation model, yielding patch features $\mathbf{X} = \{\mathbf{x}_1, \mathbf{x}_2, \ldots, \mathbf{x}_N\} \in \mathbb{R}^{N \times d}$. Each WSI is associated with an organ label $o \in \mathcal{O} = \{o_1, \ldots, o_C\}$, where $C$ is the total number of organ categories and the training set follows a long-tail distribution across organs. The goal is to generate an $L$-length diagnostically accurate report $\mathbf{Y} = \{\mathbf{y}_l\}_{l=1}^{L}$.

\subsection{Visual-Prototype Anchored Bottleneck}
\label{sec:vib} 

A WSI typically contains thousands of patches, many of which are redundant or diagnostically irrelevant, particularly diluting the discriminative information of tail-class organs. Inspired by the information bottleneck (IB) principle~\cite{IB}, we propose a Visual-Prototype Anchored Bottleneck (VPAB) that compresses redundant patch features while preserving diagnostically critical information, where prototypes clustered from labeled multi-organ WSI data serve as semantic anchors to guide the compression toward discriminative tissue patterns.


\noindent\underline{\textbf{I. Learnable Diagnostic Prototypes:}} Directly applying the Variational Information Bottleneck~\cite{variableIB} to WSIs is intractable, as each WSI constitutes a bag of thousands of patch instances and modeling the full posterior $p(z|\mathbf{x})$ over such high-dimensional bags is computationally prohibitive~\cite{pib}. We introduce a set of $K$ learnable diagnostic prototypes $\mathbf{P} = \{\mathcal{N}(\hat{z};\, \boldsymbol{\mu}_k, \boldsymbol{\Sigma}_k)\}_{k=1}^{K}$ to approximate the bag-level distribution and convert the intractable IB optimization into a tractable prototype-guided selection process. The prototype means $\{\boldsymbol{\mu}_k\}$ and covariances $\{\boldsymbol{\Sigma}_k\}$ are both learnable parameters. To provide semantically meaningful supervision, we extract region-of-interest patch-level features across diverse organs. Then, $k$-means clustering is performed over the resulting feature pool to obtain $K$ centroids $\{\mathbf{c}_k\}_{k=1}^{K}$, which capture representative tissue patterns in the histopathological feature space.

The \emph{compression} term $I(Z, X)$ in the IB framework is realized by regularizing each prototype toward its domain-informed prior $\mathcal{N}(\mathbf{c}_k, \mathbf{I})$ via KL divergence:
\begin{equation}
  \mathcal{L}_{\text{KL}} = \sum_{k=1}^{K} 
  \mathrm{KL}\big[\mathcal{N}(\boldsymbol{\mu}_k, \boldsymbol{\Sigma}_k) 
  \,\|\, \mathcal{N}(\mathbf{c}_k, \mathbf{I})\big].
\end{equation}

The \emph{predictive} term $I(Z, Y)$ is fulfilled by aligning prototype samples with clinical report semantics. We sample $\hat{\mathbf{z}}_k \sim \mathcal{N}(\boldsymbol{\mu}_k, \boldsymbol{\Sigma}_k)$ via Monte Carlo sampling and project it through an MLP layer to match the dimension of the text embedding space, then maximize its similarity with the corresponding organ report embedding $\mathbf{e}_{Y}$ from report ground truth using a pre-trained text encoder :
\begin{equation}
  \mathcal{L}_{\text{report}} = -\frac{1}{K}\sum_{k=1}^{K} 
  \mathrm{sim}(\hat{\mathbf{z}}_k,\, \mathbf{e}_{Y}),
\end{equation}
where $\mathrm{sim}(\cdot, \cdot)$ denotes cosine similarity. 

The overall Visual-Prototype Anchored Bottleneck objective combines the two terms as: $\mathcal{L}_{\text{VPAB}} = \mathcal{L}_{\text{report}} + 
\beta\,\mathcal{L}_{\text{KL}}$, where $\beta$ is a trade-off hyperparameter controlling the compression--prediction trade-off. Minimizing $\mathcal{L}_{\text{VPAB}}$ encourages the prototypes to retain maximal diagnostic information from clinical reports while compressing task-irrelevant variations toward priors.

\noindent\underline{\textbf{II. Prototype-Guided Patch Selection:}} Given patch features from a WSI $\mathbf{X} = \{\mathbf{x}_1, \mathbf{x}_2, \ldots, \mathbf{x}_N\} \in \mathbb{R}^{N \times d}$, we leverage the learned prototypes to identify diagnostically relevant patches. Specifically, we sample $\hat{\mathbf{z}}_k \sim \mathcal{N}(\boldsymbol{\mu}_k, \boldsymbol{\Sigma}_k)$ and compute the relevance score of each patch by max-pooling its cosine similarity over all prototypes:
\begin{equation}
  r_i = \max_{k \in \{1,\ldots,K\}} 
  \mathrm{sim}(\mathbf{x}_i,\, \hat{\mathbf{z}}_k).
\end{equation}
We rank all patches by their relevance scores $\{r_i\}_{i=1}^{N}$ and retain the top-$m$ ($m = \rho \cdot N$, $\rho \in (0,1)$) to form the filtered representation $\mathbf{X}_{\text{filter}} \in \mathbb{R}^{m \times d}$. This yields a compact, diagnostically informative bag representation where head-biased redundant patches are suppressed.

\subsection{Meta-Report Anchored Bank}
\label{sec:MRAB} 

Pathology reports within the same organ type naturally share recurrent linguistic patterns and standardized terminologies. We exploit this by constructing an organ-specific meta-report anchored bank and retrieving organ-faithful textual priors to steer the decoder away from head-class narrative patterns.

\noindent\underline{\textbf{I. Meta-Report Anchored Bank Construction:}} To provide organ-faithful textual priors, we construct an organ-specific meta-report anchored bank from the training split of the dataset. For each organ type $o$, we first prompt a powerful LLM (Gemini-3 Pro) to extract $M_o$ candidate meta-report templates from the training reports. To ensure representativeness, we compute the similarity between each candidate and the original reports from both biomedical-semantic (BioSimCSE)~\cite{biosimcse} and general-semantic (BGE)~\cite{bge} perspectives, and fuse the two scores into a unified ranking. Only the top-$M_o$ candidates with the highest fused scores are retained as meta-report templates. We further manually verify the selected templates for each organ to ensure clinical accuracy. The construction procedure and the meta report are provided in our code to ensure reproducibility. The resulting knowledge bank is denoted as $\mathbf{B} = \{\mathbf{b}_o\}_{o=1}^{C}$, where $\mathbf{b}_o = \{t_1^o, \ldots, t_{M_o}^o\}$. Each template is encoded by the PLIP~\cite{plip} text encoder, yielding the organ-specific feature bank $\mathbf{F} = \{\mathbf{f}_o\}_{o=1}^{C}$ with $\mathbf{f}_o \in \mathbb{R}^{M_o \times d}$.

\noindent\underline{\textbf{II. Knowledge Retrieval:}} To ensure cross-modal compatibility with the PLIP-encoded meta-reports, we re-extract patch features using the PLIP image encoder and apply the VPAB filtering indices to obtain $\mathbf{X}_{\text{filter}}^{\text{PLIP}} \in \mathbb{R}^{m \times d}$. Then we leverage the cross-attention scores from the first decoder layer to further rank selected patches by diagnostic informativeness, retaining the top-$\rho'$ proportion ($m' = \lfloor \rho' \cdot m \rfloor$) as the query set $\mathbf{Q} = \{\mathbf{q}_i\}_{i=1}^{m'} \in \mathbb{R}^{m' \times d}$. The cosine similarity between each query $\mathbf{q}_i$ and the meta-report features $\mathbf{f}_o$ is computed, and max-pooled across queries to score each template. The top-$v$ templates are selected as the retrieved meta-reports $\mathbf{F}_{\text{filter}} \in \mathbb{R}^{v \times d}$.

\subsection{Report Generation}
\label{sec:decoder}

Given the filtered visual features $\mathbf{X}_{\text{filter}}$ from VPAB and the retrieved meta-report features $\mathbf{F}_{\text{filter}}$ from MRAB, we adopt the bi-modal encoder~\cite{BiGen} to further compress both modalities. Specifically, a learnable visual token $\mathbf{h}_v$ and a learnable textual token $\mathbf{h}_t$ iteratively attend to $\mathbf{X}_{\text{filter}}$ and $\mathbf{F}_{\text{filter}}$ respectively via cross-attention layers, yielding a compact joint representation $\mathbf{H} = [\mathbf{h}_v; \mathbf{h}_t] \in \mathbb{R}^{2 \times d}$. The decoder is trained with the negative log-likelihood loss $\mathcal{L}_{\text{NLL}} = -\sum_{l=1}^{L} \log P(\mathbf{y}_l \mid \{\mathbf{y}_i\}_{i<l},\, \mathbf{H})$, where $\{\mathbf{y}_i\}_{i<l}$ denotes the previously generated tokens. Combined with the VPAB loss defined in Sec.~\ref{sec:vib}, the overall training objective is $\mathcal{L}_{\text{total}} = \mathcal{L}_{\text{NLL}} + \alpha\,\mathcal{L}_{\text{VPAB}}$, where $\alpha$ is a balancing hyperparameter.

\section{Experiments and Results}

\subsection{Implementation Details}

\noindent \underline{\textbf{I. Datasets:}} We construct a Multi-organ Long-tailed Pathology report generation dataset (ML-Path) by selecting 11 cancer types from The Cancer Genome Atlas (TCGA) based on sample frequency. The resulting dataset comprises 4,686 WSI-report pairs, naturally exhibiting a long-tailed distribution. Specifically, we categorize the cancer types into three groups: Head types including BRCA, LUNG, and KIDNEY with 879–977 samples each, Medium types including THCA, COAD, STAD, and LIHC with 365–504 samples each, and Tail types including MESO, ACC, DLBC, and CHOL with 38–60 samples each. The corresponding ground-truth reports are sourced from Chen et al. \cite{wsicaption}. We randomly split the dataset into training, validation, and testing sets, and ensure no patient-level overlap across splits and long-tail property in all sets. The split can be found in our code.


\noindent \underline{\textbf{II. Model Settings:}} Each WSI is partitioned into non-overlapping 256×256 tissue patches at 10× magnification using CLAM \cite{CLAM}, and patch-level features are extracted by UNI \cite{uni}.
The number of prototypes is set to $K=4$, and the filtering ratios $\rho$ and $\rho'$ are set to 0.4. The number of retrieved meta-report templates $v$ is set to 2. The loss balancing coefficients $\alpha$ and $\beta$ are both set to 0.1. We use the Adam optimizer with a learning rate of 1e-4, weight decay of 5e-5, and beam search with a beam size of 3. All experiments are conducted on a single NVIDIA L40-48GB GPU.

\begin{table*}[t]
\centering
\caption{Results on long-tailed dataset across 11 organ types.
B-M: BLEU-Mean (average of BLEU-1 to BLEU-4); MTR: METEOR; R-L: ROUGE-L.
\textbf{Bold}: best; \underline{Underline}: second best.
In the Mean column, shaded cells denote significant improvements over the baseline
(paired $t$-test, $n{=}11$ organs):
\colorbox{sigAAA}{$p<0.001$},
\colorbox{sigAA}{$p<0.01$},
\colorbox{sigA}{$p<0.05$};
no shading: $p \geq 0.05$.}
\label{tab:11tasks}
\resizebox{\textwidth}{!}{%
\begin{tabular}{l|ccc|ccc|ccc|ccc|ccc|ccc}
\toprule
\multirow{2}{*}{\textbf{Method}} & 
\multicolumn{3}{c|}{\textbf{BRCA}} & 
\multicolumn{3}{c|}{\textbf{LUNG}} & 
\multicolumn{3}{c|}{\textbf{KIDNEY}} & 
\multicolumn{3}{c|}{\textbf{THCA}} & 
\multicolumn{3}{c|}{\textbf{COAD}} & 
\multicolumn{3}{c}{\textbf{STAD}} \\
\cline{2-19}
 & {\scriptsize B-M} & {\scriptsize MTR} & {\scriptsize R-L}
 & {\scriptsize B-M} & {\scriptsize MTR} & {\scriptsize R-L}
 & {\scriptsize B-M} & {\scriptsize MTR} & {\scriptsize R-L}
 & {\scriptsize B-M} & {\scriptsize MTR} & {\scriptsize R-L}
 & {\scriptsize B-M} & {\scriptsize MTR} & {\scriptsize R-L}
 & {\scriptsize B-M} & {\scriptsize MTR} & {\scriptsize R-L} \\
\midrule
CNN-RNN\cite{cnn-rnn}         & 0.190 & 0.146 & 0.238 & 0.208 & 0.161 & 0.255 & 0.252 & 0.189 & 0.289 & 0.243 & 0.175 & 0.265 & 0.216 & \underline{0.171} & 0.260 & 0.207 & 0.153 & 0.243 \\
att-LSTM\cite{lstm}        & 0.176 & 0.137 & 0.253 & 0.225 & 0.153 & 0.288 & 0.286 & 0.195 & \underline{0.341} & 0.211 & 0.162 & 0.296 & 0.202 & 0.145 & 0.288 & 0.208 & 0.139 & 0.269 \\
Transformer\cite{transformer} & 0.180 & 0.137 & 0.261 & 0.252 & \underline{0.168} & 0.298 & 0.254 & 0.174 & 0.326 & 0.235 & 0.163 & 0.295 & 0.249 & 0.167 & \underline{0.309} & 0.221 & 0.147 & 0.284 \\
Wsicaption\cite{wsicaption}  & 0.207 & 0.150 & 0.280 & 0.239 & 0.163 & 0.295 & \underline{0.293} & 0.191 & 0.336 & 0.239 & 0.168 & 0.304 & 0.229 & 0.155 & 0.305 & 0.227 & 0.146 & 0.281 \\
HistoCap\cite{Histocap}    & 0.187 & 0.144 & 0.276 & 0.242 & \underline{0.168} & \textbf{0.302} & 0.278 & 0.190 & \textbf{0.342} & 0.232 & 0.163 & 0.304 & 0.218 & 0.153 & 0.297 & 0.241 & 0.155 & \textbf{0.301} \\
R2Gen\cite{r2gen}       & 0.224 & 0.157 & 0.279 & 0.237 & 0.164 & 0.296 & 0.233 & 0.204 & 0.322 & 0.244 & 0.165 & 0.290 & 0.242 & 0.166 & 0.306 & 0.218 & \underline{0.173} & 0.286 \\
R2GenCMN\cite{r2gencmn}    & 0.198 & 0.148 & 0.279 & 0.232 & 0.167 & \underline{0.299} & 0.279 & \textbf{0.218} & 0.338 & 0.245 & 0.172 & 0.306 & \underline{0.254} & 0.164 & 0.304 & 0.255 & \underline{0.173} & 0.296 \\
BiGen\cite{BiGen}       & \underline{0.240} & \underline{0.160} & \underline{0.287} & \underline{0.256} & \underline{0.168} & 0.296 & 0.279 & 0.201 & 0.319 & \underline{0.279} & \underline{0.182} & \underline{0.311} & 0.251 & 0.165 & 0.293 & \underline{0.277} & 0.167 & \underline{0.300} \\
\midrule
\textbf{Ours} & \textbf{0.250} & \textbf{0.167} & \textbf{0.291} & \textbf{0.269} & \textbf{0.183} & \textbf{0.302} & \textbf{0.324} & \underline{0.207} & \textbf{0.342} & \textbf{0.286} & \textbf{0.189} & \textbf{0.312} & \textbf{0.259} & \textbf{0.172} & \textbf{0.311} & \textbf{0.279} & \textbf{0.178} & 0.290 \\
\end{tabular}%
}

\resizebox{\textwidth}{!}{%
\begin{tabular}{l|ccc|ccc|ccc|ccc|ccc|ccc}
\toprule
\multirow{2}{*}{\textbf{Method}} & 
\multicolumn{3}{c|}{\textbf{LIHC}} & 
\multicolumn{3}{c|}{\textbf{MESO}} & 
\multicolumn{3}{c|}{\textbf{ACC}} & 
\multicolumn{3}{c|}{\textbf{CHOL}} & 
\multicolumn{3}{c|}{\textbf{DLBC}} & 
\multicolumn{3}{c}{\textbf{Mean}} \\
\cline{2-19}
 & {\scriptsize B-M} & {\scriptsize MTR} & {\scriptsize R-L}
 & {\scriptsize B-M} & {\scriptsize MTR} & {\scriptsize R-L}
 & {\scriptsize B-M} & {\scriptsize MTR} & {\scriptsize R-L}
 & {\scriptsize B-M} & {\scriptsize MTR} & {\scriptsize R-L}
 & {\scriptsize B-M} & {\scriptsize MTR} & {\scriptsize R-L}
 & {\scriptsize B-M} & {\scriptsize MTR} & {\scriptsize R-L} \\
\hline
CNN-RNN\cite{cnn-rnn} & 0.206 & 0.151 & 0.245 & 0.156 & 0.122 & 0.210 & 0.176 & 0.139 & 0.245 & 0.168 & \underline{0.139} & 0.216 & 0.152 & 0.092 & 0.226 & \cellcolor{sigAAA}0.215 & \cellcolor{sigAAA}0.164 & \cellcolor{sigAAA}0.250 \\
att-LSTM\cite{lstm} & \underline{0.229} & 0.158 & 0.279 & 0.163 & 0.103 & 0.241 & 0.177 & 0.109 & 0.237 & 0.156 & 0.120 & 0.242 & 0.160 & 0.084 & 0.214 & \cellcolor{sigAAA}0.221 & \cellcolor{sigAAA}0.156 & \cellcolor{sigAAA}0.287 \\
Transformer\cite{transformer} & \textbf{0.247} & \underline{0.163} & \underline{0.293} & 0.183 & 0.117 & 0.247 & 0.188 & 0.118 & 0.259 & 0.176 & 0.136 & 0.247 & 0.161 & 0.113 & 0.242 & \cellcolor{sigAAA}0.230 & \cellcolor{sigAAA}0.157 & \cellcolor{sigAA}0.293 \\
Wsicaption\cite{wsicaption} & 0.225 & 0.152 & 0.291 & 0.185 & 0.116 & 0.258 & 0.189 & 0.118 & 0.264 & \underline{0.184} & 0.130 & \underline{0.261} & 0.168 & 0.131 & 0.241 & \cellcolor{sigAAA}0.239 & \cellcolor{sigAAA}0.162 & \cellcolor{sigAA}0.298 \\
HistoCap\cite{Histocap} & 0.221 & 0.150 & 0.287 & \underline{0.200} & 0.117 & \underline{0.266} & 0.217 & 0.129 & \underline{0.282} & 0.147 & 0.120 & 0.239 & \underline{0.203} & 0.125 & \underline{0.282} & \cellcolor{sigAAA}0.231 & \cellcolor{sigAAA}0.161 & \underline{0.301} \\
R2Gen\cite{r2gen} & 0.204 & 0.153 & \textbf{0.294} & 0.150 & 0.114 & 0.232 & \textbf{0.222} & \underline{0.146} & \textbf{0.290} & 0.161 & 0.126 & 0.231 & 0.200 & \underline{0.143} & 0.246 & \cellcolor{sigAA}0.247 & \cellcolor{sigAA}0.170 & \cellcolor{sigA}0.295 \\
R2GenCMN\cite{r2gencmn} & 0.228 & 0.160 & 0.282 & 0.153 & 0.119 & 0.242 & 0.197 & 0.136 & 0.267 & 0.171 & \underline{0.139} & 0.243 & 0.145 & 0.126 & 0.261 & \cellcolor{sigAAA}0.259 & \cellcolor{sigAA}\underline{0.172} & \cellcolor{sigA}0.300 \\
BiGen\cite{BiGen} & 0.226 & 0.159 & 0.279 & 0.175 & \underline{0.131} & 0.251 & 0.202 & 0.120 & 0.266 & 0.156 & 0.123 & 0.239 & \textbf{0.230} & \underline{0.143} & \textbf{0.287} & \cellcolor{sigA}\underline{0.268} & \cellcolor{sigAAA}\underline{0.172} & 0.297 \\
\hline
\textbf{Ours} & \textbf{0.247} & \textbf{0.170} & \textbf{0.294} & \textbf{0.203} & \textbf{0.143} & \textbf{0.273} & \underline{0.220} & \textbf{0.147} & 0.268 & \textbf{0.213} & \textbf{0.150} & \textbf{0.272} & 0.200 & \textbf{0.146} & 0.273 & \textbf{0.273} & \textbf{0.181} & \textbf{0.305} \\
\Xhline{1.2pt}
\end{tabular}%
}
\
\end{table*}

\subsection{Multi-Organ Report Generation Results}

For all comparative methods, we use exactly the same data split to ensure fairness of the experiments. We compare our method with eight existing approaches in Table \ref{tab:11tasks}, including two LSTM-based methods: CNN-RNN \cite{cnn-rnn} and att-LSTM \cite{lstm}, and six Transformer-based methods: Transformer \cite{transformer}, R2Gen \cite{r2gen} and R2GenCMN \cite{r2gencmn} (both originally designed for radiology report generation), Wsicaption \cite{wsicaption} and HistoCap \cite{Histocap} (designed for pathology report generation), and BiGen \cite{BiGen} which also serves as a naive retrieval baseline. Following~\cite{wsicaption,BiGen}, three widely used Natural Language Processing metrics are adopted: BLEU-Mean (the average of BLEU-1 to BLEU-4), METEOR, and ROUGE-L.

As shown in Table \ref{tab:11tasks}, our method achieves the best overall Mean scores across all three metrics. Existing methods such as BiGen~\cite{BiGen} perform competitively on head-class organs but degrade notably on tail classes due to the lack of explicit long-tail handling. In contrast, our method yields substantial improvements on tail classes: on CHOL, we achieve 0.213 in B-M, surpassing Wsicaption~\cite{wsicaption} (0.184) by +15.8\%; on MESO, we obtain 0.203 in B-M and 0.143 in MTR, outperforming BiGen~\cite{BiGen} by +16.0\% and +9.2\%, respectively. Importantly, these tail-class gains do not compromise head-class performance, confirming that our framework effectively mitigates long-tail bias without sacrificing overall generation quality.

\textbf{Discussion.} We attribute the improvements to the complementary design of VPAB and MRAB. VPAB compensates for the scarce visual representations of tail-class organs by providing organ-specific
prototypes, while MRAB prevents the decoder from defaulting to head-class expressions by injecting organ-faithful textual priors. The two modules jointly address the long-tail bias, explaining the consistent tail-class gains without head-class degradation.

\subsection{Ablation Study}

\begin{table*}[t]
\centering
\caption{\textbf{Ablation study of different components.}}
\label{tab:ablation_results}
\resizebox{\textwidth}{!}{%
\begin{tabular}{ccc|ccc|ccc|ccc|ccc}
\toprule
\multicolumn{3}{c|}{\scriptsize\textbf{Components}} 
 & \multicolumn{3}{c|}{\scriptsize\textbf{Overall}}
 & \multicolumn{3}{c|}{\scriptsize\textbf{Head}}
 & \multicolumn{3}{c|}{\scriptsize\textbf{Medium}}
 & \multicolumn{3}{c}{\scriptsize\textbf{Tail}} \\
\cmidrule(lr){1-3} \cmidrule(lr){4-6} \cmidrule(lr){7-9} \cmidrule(lr){10-12} \cmidrule(lr){13-15}

\scriptsize{BASE} & \scriptsize{MRAB} & \scriptsize{VPAB}
 & {\scriptsize B-M} & {\scriptsize MTR} & {\scriptsize R-L}
 & {\scriptsize B-M} & {\scriptsize MTR} & {\scriptsize R-L}
 & {\scriptsize B-M} & {\scriptsize MTR} & {\scriptsize R-L}
 & {\scriptsize B-M} & {\scriptsize MTR} & {\scriptsize R-L} \\
\midrule

\cmark & \xmark & \xmark
 & \scriptsize{0.259} & \scriptsize{0.170} & \scriptsize{0.290} 
 & \scriptsize{0.266} & \scriptsize{0.176} & \scriptsize{0.297} 
 & \scriptsize{0.254} & \scriptsize{0.165} & \scriptsize{0.285} 
 & \scriptsize{0.202} & \scriptsize{0.121} & \scriptsize{0.240} \\ 

\cmark & \cmark & \xmark
 & \scriptsize{0.271} & \scriptsize{0.178} & \scriptsize{0.301} 
 & \scriptsize{0.275} & \scriptsize{0.183} & \scriptsize{0.305} 
 & \scriptsize{0.266} & \scriptsize{0.175} & \scriptsize{0.299} 
 & \scriptsize{0.212} & \scriptsize{0.127} & \scriptsize{0.250} \\ 

\cmark & \xmark & \cmark
 & \scriptsize{0.268} & \scriptsize{0.174} & \scriptsize{0.295} 
 & \scriptsize{0.276} & \scriptsize{0.181} & \scriptsize{0.298} 
 & \scriptsize{0.261} & \scriptsize{0.171} & \scriptsize{0.296} 
 & \scriptsize{0.211} & \scriptsize{0.132} & \scriptsize{0.245} \\ 

\cmark & \cmark & \cmark
 & \scriptsize\textbf{0.273} & \scriptsize\textbf{0.181} & \scriptsize\textbf{0.305}
 & \scriptsize\textbf{0.278} & \scriptsize\textbf{0.186} & \scriptsize\textbf{0.308}
 & \scriptsize\textbf{0.270} & \scriptsize\textbf{0.178} & \scriptsize\textbf{0.302}
 & \scriptsize\textbf{0.220} & \scriptsize\textbf{0.141} & \scriptsize\textbf{0.264} \\

\bottomrule
\end{tabular}%
}
\end{table*}

\begin{figure}[t] 
    \centering
    \includegraphics[width=0.85\linewidth]{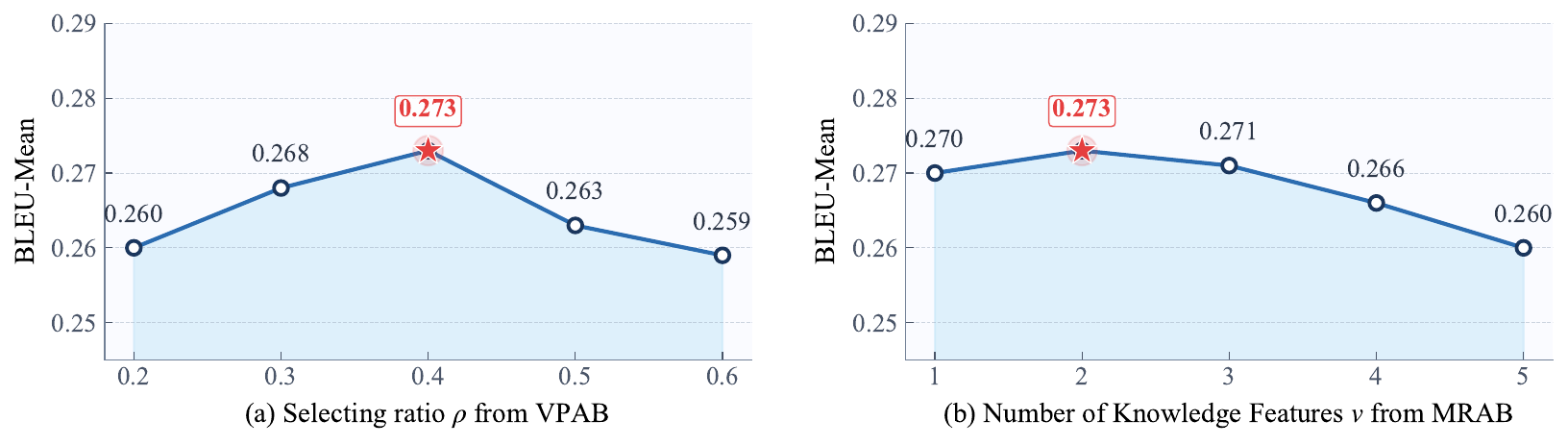} 
    \caption{Experimental results of BLEU-Mean with different parameters.}
    \label{fig3}
\end{figure}

To validate the contribution of each component, we conduct ablation experiments by integrating MRAB and VPAB into the baseline model. As shown in Table~\ref{tab:ablation_results}, the baseline alone reveals a significant Head–Tail performance gap (B-M: 0.266 vs. 0.202), confirming the severity of long-tail bias. Adding MRAB alone improves overall B-M from 0.259 to 0.271, with the Tail group gaining +4.9\% in B-M and +4.2\% in R-L, validating that organ-faithful textual priors effectively alleviate decoding bias. Incorporating VPAB alone yields an overall B-M of 0.268, with Tail MTR improving by +9.1\%, demonstrating that prototype-guided visual compression suppresses head-biased redundancy. When both modules are combined, the full model achieves the best results across all groups. The Tail group benefits most significantly (B-M: +8.9\%, MTR: +16.5\%, R-L: +10.0\% over the baseline), while Head and Medium groups also exhibit consistent gains.


Fig.~\ref{fig3} illustrates the sensitivity of two key hyperparameters: the selecting ratio $\rho$ in VPAB and the number of retrieved meta-reports $v$ in MRAB. Both exhibit a rise-then-fall pattern, peaking at $\rho=0.4$ and $v=2$, respectively. A small $\rho$ discards useful visual information while a large value retains redundancy. Similarly, excessive retrieved meta-reports introduce noise and conflicting textual priors. Additional sensitivity analysis on the number of VPAB prototypes further shows that $K=4$ yields the best performance. We thus adopt $\rho=0.4$, $v=2$, and $K=4$ as the default settings throughout all experiments.



%
%
%
\section{Conclusion}
In this paper, we present PriOrGen, a dual prior-anchored framework that addresses the long-tail distribution challenge in multi-organ pathology report generation. The proposed VPAB and MRAB modules jointly mitigate visual representation bias and textual decoding bias from complementary perspectives. Experiments on 11 organ types demonstrate consistent improvements, particularly on tail-class organs. This work establishes the first systematic study of long-tail bias in multi-organ pathology report generation and offers a general dual-prior paradigm that jointly mitigates distributional imbalance from both visual and textual perspectives, with potential extensions to other long-tailed medical report generation scenarios.

\par\medskip
\noindent\textbf{Acknowledgements.} This work was supported by the Institute of Digital Medicine, City University of Hong Kong, and the Research Grants Council of the Hong Kong Special Administrative Region, China [Project No. CityU11208324].


\bibliographystyle{splncs04}
\bibliography{mybib}

@article{zongshu1,
  title={From whole-slide image to biomarker prediction: end-to-end weakly supervised deep learning in computational pathology},
  author={El Nahhas, Omar SM and van Treeck, Marko and W{\"o}lflein, Georg and Unger, Michaela and Ligero, Marta and Lenz, Tim and Wagner, Sophia J and Hewitt, Katherine J and Khader, Firas and Foersch, Sebastian and others},
  journal={Nature protocols},
  volume={20},
  number={1},
  pages={293--316},
  year={2025},
  publisher={Nature Publishing Group UK London}
}

@article{CLAM,
  title={Data-efficient and weakly supervised computational pathology on whole-slide images},
  author={Lu, Ming Y and Williamson, Drew FK and Chen, Tiffany Y and Chen, Richard J and Barbieri, Matteo and Mahmood, Faisal},
  journal={Nature biomedical engineering},
  volume={5},
  number={6},
  pages={555--570},
  year={2021},
  publisher={Nature Publishing Group UK London}
}

@article{zongshu3,
  title={Clinical-grade computational pathology using weakly supervised deep learning on whole slide images},
  author={Campanella, Gabriele and Hanna, Matthew G and Geneslaw, Luke and Miraflor, Allen and Werneck Krauss Silva, Vitor and Busam, Klaus J and Brogi, Edi and Reuter, Victor E and Klimstra, David S and Fuchs, Thomas J},
  journal={Nature medicine},
  volume={25},
  number={8},
  pages={1301--1309},
  year={2019},
  publisher={Nature Publishing Group US New York}
}

@inproceedings{wsicaption,
  title={Wsicaption: Multiple instance generation of pathology reports for gigapixel whole-slide images},
  author={Chen, Pingyi and Li, Honglin and Zhu, Chenglu and Zheng, Sunyi and Shui, Zhongyi and Yang, Lin},
  booktitle={International Conference on Medical Image Computing and Computer-Assisted Intervention},
  pages={546--556},
  year={2024},
  organization={Springer}
}

@inproceedings{histgen,
  title={Histgen: Histopathology report generation via local-global feature encoding and cross-modal context interaction},
  author={Guo, Zhengrui and Ma, Jiabo and Xu, Yingxue and Wang, Yihui and Wang, Liansheng and Chen, Hao},
  booktitle={International Conference on Medical Image Computing and Computer-Assisted Intervention},
  pages={189--199},
  year={2024},
  organization={Springer}
}

@article{tran2024generating,
  title={Generating highly accurate pathology reports from gigapixel whole slide images with HistoGPT},
  author={Tran, Manuel and Schmidle, Paul and Wagner, Sophia J and Koch, Valentin and Lupperger, Valerio and Feuchtinger, Annette and B{\"o}hner, Alexander and Kaczmarczyk, Robert and Biedermann, Tilo and Eyerich, Kilian and others},
  journal={Medrxiv},
  pages={2024--03},
  year={2024},
  publisher={Cold Spring Harbor Laboratory Press}
}

@inproceedings{BiGen,
  title={Historical report guided bi-modal concurrent learning for pathology report generation},
  author={Zhang, Ling and Yun, Boxiang and Li, Qingli and Wang, Yan},
  booktitle={International Conference on Medical Image Computing and Computer-Assisted Intervention},
  pages={343--352},
  year={2025},
  organization={Springer}
}

@article{Long-Tai-cls1,
  title={MONICA: Benchmarking on Long-tailed Medical Image Classification},
  author={Ju, Lie and Yan, Siyuan and Zhou, Yukun and Nan, Yang and Xing, Xiaodan and Duan, Peibo and Ge, Zongyuan},
  journal={arXiv preprint arXiv:2410.02010
        
        
        
        
        
        
        
        
        
        
        
        
        
        },
  year={2024}
}

@article{Long-Tai-cls2,
  title={Long-tailed medical diagnosis with relation-aware representation learning and iterative classifier calibration},
  author={Pan, Li and Zhang, Yupei and Yang, Qiushi and Li, Tan and Chen, Zhen},
  journal={Computers in Biology and Medicine},
  volume={188},
  pages={109772},
  year={2025},
  publisher={Elsevier}
}

@inproceedings{Long-Tai-cls3,
  title={Bpaco: Balanced parametric contrastive learning for long-tailed medical image classification},
  author={Cai, Zhiyuan and Wei, Tianyunxi and Lin, Li and Chen, Hao and Tang, Xiaoying},
  booktitle={International Conference on Medical Image Computing and Computer-Assisted Intervention},
  pages={383--393},
  year={2024},
  organization={Springer}
}

@inproceedings{Long-Tai-cls4,
  title={Combat long-tails in medical classification with relation-aware consistency and virtual features compensation},
  author={Pan, Li and Zhang, Yupei and Yang, Qiushi and Li, Tan and Chen, Zhen},
  booktitle={International Conference on Medical Image Computing and Computer-Assisted Intervention},
  pages={14--23},
  year={2023},
  organization={Springer}
}

@article{uni,
  title={Towards a general-purpose foundation model for computational pathology},
  author={Chen, Richard J and Ding, Tong and Lu, Ming Y and Williamson, Drew FK and Jaume, Guillaume and Song, Andrew H and Chen, Bowen and Zhang, Andrew and Shao, Daniel and Shaban, Muhammad and others},
  journal={Nature medicine},
  volume={30},
  number={3},
  pages={850--862},
  year={2024},
  publisher={Nature Publishing Group US New York}
}

@inproceedings{cnn-rnn,
  title={Show and tell: A neural image caption generator},
  author={Vinyals, Oriol and Toshev, Alexander and Bengio, Samy and Erhan, Dumitru},
  booktitle={Proceedings of the IEEE conference on computer vision and pattern recognition},
  pages={3156--3164},
  year={2015}
}

@inproceedings{lstm,
  title={Show, attend and tell: Neural image caption generation with visual attention},
  author={Xu, Kelvin and Ba, Jimmy and Kiros, Ryan and Cho, Kyunghyun and Courville, Aaron and Salakhudinov, Ruslan and Zemel, Rich and Bengio, Yoshua},
  booktitle={International conference on machine learning},
  pages={2048--2057},
  year={2015},
  organization={PMLR}
}

@article{transformer,
  title={Attention is all you need},
  author={Vaswani, Ashish and Shazeer, Noam and Parmar, Niki and Uszkoreit, Jakob and Jones, Llion and Gomez, Aidan N and Kaiser, {\L}ukasz and Polosukhin, Illia},
  journal={Advances in neural information processing systems},
  volume={30},
  year={2017}
}

@inproceedings{r2gen,
  title={Generating radiology reports via memory-driven transformer},
  author={Chen, Zhihong and Song, Yan and Chang, Tsung-Hui and Wan, Xiang},
  booktitle={Proceedings of the 2020 conference on empirical methods in natural language processing (EMNLP)},
  pages={1439--1449},
  year={2020}
}

@inproceedings{r2gencmn,
  title={Cross-modal memory networks for radiology report generation},
  author={Chen, Zhihong and Shen, Yaling and Song, Yan and Wan, Xiang},
  booktitle={Proceedings of the 59th annual meeting of the association for computational linguistics and the 11th international joint conference on natural language processing (volume 1: long papers)},
  pages={5904--5914},
  year={2021}
}

@inproceedings{Histocap,
  title={Automatic report generation for histopathology images using pre-trained vision transformers and BERT},
  author={Sengupta, Saurav and Brown, Donald E},
  booktitle={2024 IEEE International Symposium on Biomedical Imaging (ISBI)},
  pages={1--5},
  year={2024},
  organization={IEEE}
}

@article{pib,
  title={Prototypical information bottlenecking and disentangling for multimodal cancer survival prediction},
  author={Zhang, Yilan and Xu, Yingxue and Chen, Jianqi and Xie, Fengying and Chen, Hao},
  journal={arXiv preprint arXiv:2401.01646
        
        
        
        
        
        
        
        
        
        
        
        
        
         },
  year={2024}
}

@article{IB,
  title={The information bottleneck method},
  author={Tishby, Naftali and Pereira, Fernando C and Bialek, William},
  journal={arXiv preprint physics/0004057},
  year={2000}
}

@article{variableIB,
  title={Deep variational information bottleneck},
  author={Alemi, Alexander A and Fischer, Ian and Dillon, Joshua V and Murphy, Kevin},
  journal={arXiv preprint arXiv:1612.00410
        
        
        
        
        },
  year={2016}
}

@article{plip,
  title={A visual--language foundation model for pathology image analysis using medical twitter},
  author={Huang, Zhi and Bianchi, Federico and Yuksekgonul, Mert and Montine, Thomas J and Zou, James},
  journal={Nature medicine},
  volume={29},
  number={9},
  pages={2307--2316},
  year={2023},
  publisher={Nature Publishing Group US New York}
}

@article{longtail1,
  title={Decoupling representation and classifier for long-tailed recognition},
  author={Kang, Bingyi and Xie, Saining and Rohrbach, Marcus and Yan, Zhicheng and Gordo, Albert and Feng, Jiashi and Kalantidis, Yannis},
  journal={arXiv preprint arXiv:1910.09217},
  year={2019}
}

@inproceedings{longtail2,
  title={Improving factual completeness and consistency of image-to-text radiology report generation},
  author={Miura, Yasuhide and Zhang, Yuhao and Tsai, Emily and Langlotz, Curtis and Jurafsky, Dan},
  booktitle={Proceedings of the 2021 Conference of the North American Chapter of the Association for Computational Linguistics: Human Language Technologies},
  pages={5288--5304},
  year={2021}
}

@article{longtail3,
  title={Adjusting logit in Gaussian form for long-tailed visual recognition},
  author={Li, Mengke and Cheung, Yiu-ming and Lu, Yang and Hu, Zhikai and Lan, Weichao and Huang, Hui},
  journal={IEEE Transactions on Artificial Intelligence},
  volume={5},
  number={10},
  pages={5026--5039},
  year={2024},
  publisher={IEEE}
}

@inproceedings{biosimcse,
  title={Biosimcse: Biomedical sentence embeddings using contrastive learning},
  author={Kanakarajan, Kamal Raj and Kundumani, Bhuvana and Abraham, Abhijith and Sankarasubbu, Malaikannan},
  booktitle={Proceedings of the 13th international workshop on health text mining and information analysis (LOUHI)},
  pages={81--86},
  year={2022}
}

@inproceedings{bge,
  title={C-pack: Packed resources for general chinese embeddings},
  author={Xiao, Shitao and Liu, Zheng and Zhang, Peitian and Muennighoff, Niklas and Lian, Defu and Nie, Jian-Yun},
  booktitle={Proceedings of the 47th international ACM SIGIR conference on research and development in information retrieval},
  pages={641--649},
  year={2024}
}
\end{document}